\documentclass[sigconf]{acmart}
\AtBeginDocument{%
  }

\usepackage{booktabs}
\usepackage{multirow}
\usepackage{amsmath}
\usepackage{enumitem}
\usepackage{graphicx}
\usepackage{float}
\usepackage{placeins}
\usepackage[linesnumbered,ruled,vlined]{algorithm2e}

\setcopyright{cc}
\setcctype{by}
\copyrightyear{2026}
\acmYear{2026}
\acmDOI{10.1145/3767308.3836597}
\acmConference[MM '26]{Proceedings of the 34th ACM International Conference on Multimedia}{November 10--14, 2026}{Rio de Janeiro, Brazil}
\acmBooktitle{Proceedings of the 34th ACM International Conference on Multimedia (MM '26), November 10--14, 2026, Rio de Janeiro, Brazil}
\acmISBN{979-8-4007-2213-4/2026/11}

\begin{document}

\title{Discovering Diverse Planning Policies for Multimodal Embodied Agents with Quality-Diversity Optimization}

\author{Pengfei Xu}
\affiliation{%
  \institution{School of Artificial Intelligence, Nanjing University of Information Science and Technology}
  \city{Nanjing}
  \state{Jiangsu}
  \country{China}}
\email{202512492840@nuist.edu.cn}

\author{Yong Liu}
\affiliation{%
  \institution{School of Artificial Intelligence, Nanjing University of Information Science and Technology}
  \city{Nanjing}
  \state{Jiangsu}
  \country{China}}
\email{202412491463@nuist.edu.cn}

\author{Xiaoya Nan}
\affiliation{%
  \institution{School of Artificial Intelligence, Nanjing University of Information Science and Technology}
  \city{Nanjing}
  \state{Jiangsu}
  \country{China}}
\email{xynan@nuist.edu.cn}

\author{Qiang Yang}
\affiliation{%
  \institution{School of Artificial Intelligence, Nanjing University of Information Science and Technology}
  \city{Nanjing}
  \state{Jiangsu}
  \country{China}}
\email{qiang_yang@nuist.edu.cn}

\author{Peilan Xu}
\authornote{Corresponding author.}
\affiliation{%
  \institution{School of Artificial Intelligence, Nanjing University of Information Science and Technology}
  \city{Nanjing}
  \state{Jiangsu}
  \country{China}}
\email{xpl@nuist.edu.cn}

\renewcommand{\shortauthors}{Xu et al.}

\begin{abstract}
Multimodal embodied agents are increasingly required to solve long-horizon tasks by integrating visual observations, textual goals, and interaction history into closed-loop decision making. However, state-of-the-art large-model-based planners often rely on a single dominant planning style during execution. Once this execution mode becomes ineffective, the agent may remain stalled for many steps, repeatedly interacting with the environment without making meaningful progress. We address this limitation by proposing a Quality-Diversity (QD) framework for discovering diverse planning policies for multimodal embodied agents. The proposed method treats planning-policy templates as evolvable individuals and organizes them into a behavior-indexed archive rather than collapsing search to a single prompt style. In the offline stage, rollout trajectories are summarized into structured success and failure experiences, which guide policy variation through recombination and experience-guided mutation. The resulting policies are mapped into a behavior space defined by interaction intensity and goal-directedness, and the highest-quality policy in each niche is retained in the archive. In the online stage, the agent executes one policy at a time while monitoring task progress. When persistent stall is detected, the system rolls back to the latest checkpoint and switches to a behaviorally distinct archive policy to resume execution. Experiments on the ThreeDWorld transport benchmark show that the proposed framework improves both task success and interaction efficiency over representative baseline planners. These results suggest that discovering diverse policy repertoires is an effective way to support adaptive multimodal planning and online failure recovery. Code is available at https://github.com/EvoNexusX/2026XuQD-Plan.git.
\end{abstract}



\maketitle
\thispagestyle{plain}

\begin{figure}[t!]
	\centering
	\includegraphics[width=\columnwidth]{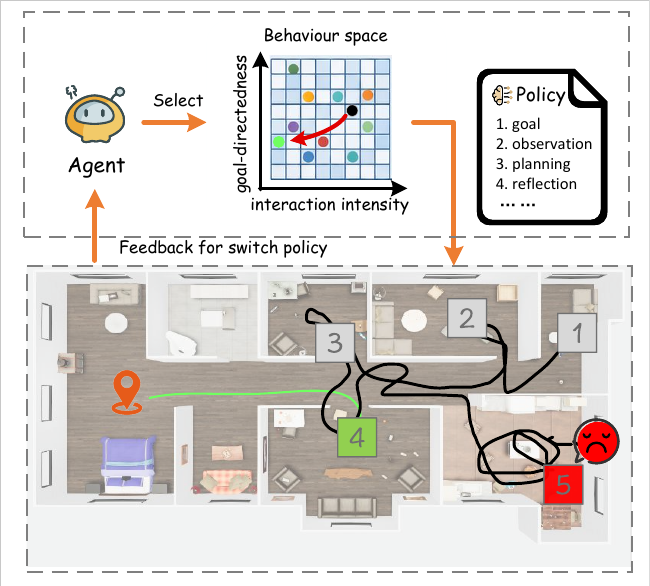}
	\caption{Motivation of archive-based behavioral recovery. The agent first retrieves a planning-policy template from a behavior archive and follows it through intermediate states 1--5. When the current policy becomes stalled, with repetitive looping shown here as one representative failure pattern, the system rolls back to the latest recoverable state 4 and switches to a behaviorally distinct policy. The new policy enables the agent to resume progress instead of repeating the previous unproductive trajectory.}
	\Description{Motivation figure: archive-based recovery with rollback and policy switching.}
	\label{fig:intro_motivation}
\end{figure}

\section{Introduction}

Interactive multimodal agents are increasingly required to perceive, reason, and act in partially observable environments using visual observations, textual goals, and execution history. This challenge is particularly evident in long-horizon embodied tasks, where an agent must repeatedly interpret multimodal feedback and convert it into coherent action sequences over extended interaction horizons~\cite{gao2025building,song2023llm_planner,chen2025exploring}. Although large language models (LLMs) and multimodal large language models (MLLMs) have shown strong zero-shot reasoning ability, their deployment in closed-loop embodied environments still faces a practical limitation: a \emph{single} planning style can become brittle when persistent feedback indicates that the agent should revise \emph{how} it is acting, rather than simply continue the current plan.

A central difficulty in long-horizon multimodal interaction is that failure often does not come from a single catastrophic action. Instead, the agent may execute a sequence of locally plausible decisions that yields little overall progress. Once execution becomes locked into an unsuitable behavioral mode, the agent may continue interacting with the environment without advancing the task, even though visual and environmental feedback remains available. Figure~\ref{fig:intro_motivation} illustrates this phenomenon. The agent first follows a planning-policy template retrieved from a behavior archive, becomes stalled near state 5, and resumes progress only after rollback and policy switching. This observation suggests that the bottleneck is not only whether the agent can reason at each step, but also whether it can \emph{change behavioral mode} when its current interaction style is no longer effective.

Existing long-horizon agent methods have substantially improved reasoning--action coupling, deliberate search, and feedback-guided decision making, as exemplified by ReAct~\cite{yao2022react}, Tree-of-Thoughts~\cite{yao2023tree}, and Reflexion~\cite{shinn2023reflexion}. Embodied multimodal systems have further strengthened grounding and execution through richer perceptual inputs and iterative interaction mechanisms~\cite{huang2023inner,gao2024fast}. However, these methods still typically rely on one dominant planning style during deployment, even when their internal reasoning procedures are sophisticated. As a result, they are usually better at \emph{executing} a strategy than at \emph{switching} to a different strategy once the current mode becomes unproductive in a specific multimodal context.

To address this limitation, we propose an archive-based planning framework that treats prompt templates as \emph{planning-policy templates}, namely reusable carriers of high-level behavioral style. Rather than searching for a single best prompt, the proposed method constructs an explicit archive of behaviorally diverse planning policies and uses that archive as a recovery repertoire during online execution. Offline, the method discovers complementary planning behaviors from multimodal rollout trajectories through Quality-Diversity (QD) search. Online, a stall detector monitors task progress and triggers switching to a behaviorally distinct archive entry when the current policy becomes unproductive. In this way, behavioral diversity becomes a practical mechanism for robust multimodal planning and online recovery. We evaluate the method on the ThreeDWorld (TDW) transport benchmark and show that archive-based adaptation improves both task success and interaction efficiency under a strict interaction budget.

\noindent\textbf{Contributions.}
This work makes three contributions. First, it identifies \emph{behavioral lock-in} under static planning styles as a major source of brittleness in long-horizon multimodal embodied tasks, and formulates recovery as a behavior-level adaptation problem. Second, it introduces an archive-based planning-policy discovery framework that uses QD search to organize behaviorally diverse multimodal planning strategies rather than collapsing search to a single prompt style. Third, it develops an online adaptive recovery mechanism based on checkpointing, stall detection, and behavior-aware policy switching, and demonstrates that this combination improves both task completion and interaction efficiency on TDW.

\section{Related Work}
\label{sec:related}

\subsection{Multimodal Planning with Large Models}
Large-model-based embodied agents have significantly expanded the role of language and multimodal reasoning in interactive decision making. Early systems such as SayCan~\cite{ahn2022can} and ALFWorld~\cite{shridharalfworld} connected language models to executable action spaces, enabling goal decomposition and action sequencing in embodied environments. More recent works have strengthened multimodal grounding and long-horizon planning. RT-2~\cite{zitkovich2023rt} emphasizes the transfer from vision-language representations to robotic control, while LLM-Planner~\cite{song2023llm_planner} highlights explicit planning structures for long-horizon tasks. Embodied EvoAgent~\cite{gao2025building} further shows that planning quality can be improved through iterative interaction and refinement. These studies demonstrate the growing capacity of large models to support multimodal embodied planning, but they mainly focus on improving planning competence within a single execution style.

\subsection{Failure Recovery and Online Adaptation in Long-Horizon Agents}
A related line of research studies how long-horizon agents recover from execution errors and adapt online. Reflexion~\cite{shinn2023reflexion} uses verbal reinforcement to revise future behavior based on previous failures, while inner-monologue-style agents~\cite{huang2023inner} incorporate intermediate self-feedback into the decision process. Other methods focus on test-time or online adaptation using newly observed environmental information~\cite{gao2024fast}. Planning-oriented approaches such as ReAct~\cite{yao2022react} and Tree-of-Thoughts~\cite{yao2023tree} also alleviate execution drift by improving intermediate reasoning and exploration. Despite these advances, most existing methods still aim to repair or refine the \emph{current} planning process, rather than explicitly switching to an alternative behavioral policy when the current execution mode becomes persistently ineffective.

\subsection{Quality-Diversity and Prompt-Level Policy Search}
Quality-Diversity (QD) algorithms seek a diverse set of high performing solutions rather than a single optimum. MAP-Elites~\cite{mouret2015illuminating} is a representative framework in this line and has inspired extensive work on diversity-aware search~\cite{lehman2011abandoning,pugh2016qualitydiversity}. QD has shown particular value in domains such as evolutionary robotics and control, where structured behavioral repertoires improve robustness under changing conditions. In parallel, prompt optimization methods such as EvoPrompt~\cite{guo2023evoprompt} and Promptbreeder~\cite{fernando2024promptbreeder} demonstrate that prompt text itself can be optimized and evolved. Our work connects these two directions in a multimodal embodied setting. Instead of searching for one best prompt, we use QD to construct a \emph{behavior archive} of planning-policy templates and use that archive for online recovery under multimodal execution feedback.

\section{Method}
\label{sec:method}

\begin{figure*}[htbp]
	\centering
	\includegraphics[width=0.85\textwidth]{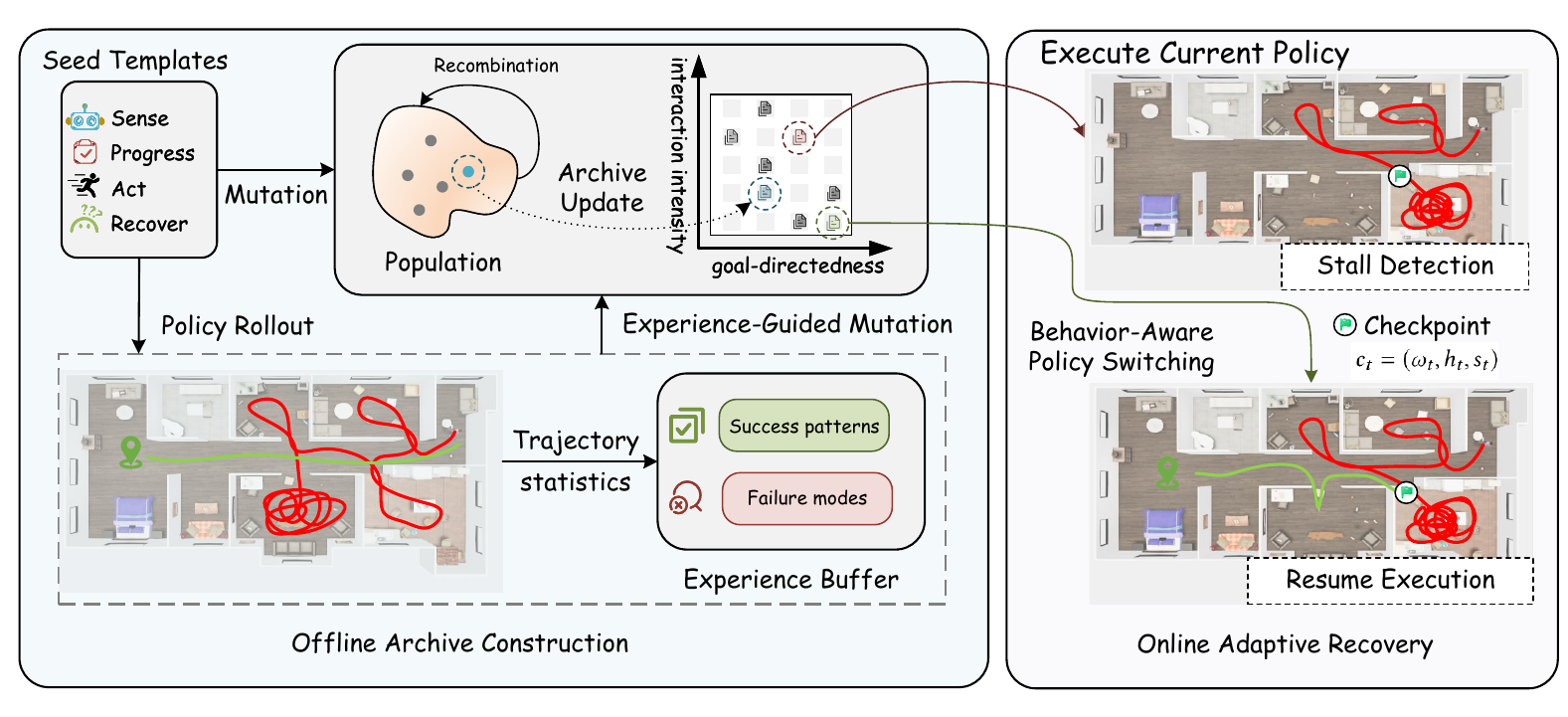}
	\caption{Overview of the proposed framework. The method contains an offline archive-construction stage and an online adaptive-recovery stage. Offline, seed planning-policy templates are evolved through recombination and experience-guided mutation. Policy rollouts produce trajectory statistics, from which success patterns and failure modes are extracted and stored in an experience buffer. Policies are mapped into a behavior space defined by \emph{interaction intensity} and \emph{goal-directedness}, and the highest-quality policy in each niche is stored in a behavior-indexed archive. Online, the agent executes one policy while monitoring task progress. When persistent stall is detected, the system rolls back to the latest checkpoint and switches to a behaviorally distinct policy retrieved from the archive. Thus, behavioral diversity discovered offline becomes a practical recovery resource during online execution.}
	\Description{Framework overview: offline archive construction and online adaptive recovery.}
	\label{fig:overview}
\end{figure*}

\subsection{Overview}
The proposed framework contains two tightly coupled stages: offline archive construction and online adaptive recovery. Offline, we search for a set of high-quality yet behaviorally distinct planning-policy templates and organize them into an archive indexed by behavior descriptors. Online, the agent executes one policy at a time in a closed loop. When the current policy fails to make sustained task progress, the agent rolls back to the latest recoverable checkpoint and switches to a behaviorally distinct archive entry. Therefore, behavioral diversity is used not only as an offline search objective but also as an executable recovery resource during long-horizon multimodal interaction.

\subsection{Problem Formulation}
We consider an embodied agent acting in a partially observable environment to solve long-horizon tasks. At each time step $t$ , the agent receives a multimodal observation $o_t \in \mathcal{O}$, maintains an interaction history $h_t$, and tracks a task-progress state $s_t \in \mathcal{S}$. The progress state summarizes the current execution status, such as milestone completion, and serves as the observable signal for online recovery.

Let $\tau \in \mathcal{T}$ denote a planning-policy template. When instantiated with the current multimodal context, $\tau$ induces an action-selection policy $\pi_\tau$ over the action space $\mathcal{A}$. Formally, for any $a \in \mathcal{A}$,
\begin{equation}
	\label{eq:policy}
	\pi_\tau(a \mid o_t,h_t,s_t)
	=
	\Pr\!\big(a \mid \phi(\tau; o_t,h_t,s_t)\big),
\end{equation}
where $\phi(\tau; o_t,h_t,s_t)$ denotes the instantiated prompt built from template $\tau$ and context $(o_t,h_t,s_t)$. Different templates may therefore induce different planning styles under similar task conditions.

Each template $\tau$ is evaluated through rollout execution and is associated with two quantities. The first is a behavior descriptor $\mathrm{BD}(\tau)$ that places the induced policy in a low-dimensional behavior space. The second is a quality score $Q(\tau)$ that measures execution performance. Based on these two quantities, we maintain an archive $\mathcal{M}$ whose cells correspond to discretized behavioral niches and whose entries store the best policy found so far in each niche.

The goal of the offline stage is to construct an archive that preserves multiple effective planning styles rather than collapsing search to a single dominant policy. The goal of the online stage is to use this archive as a recovery repertoire, so that execution can switch away from an unproductive policy when sustained stagnation is detected.

\subsection{Offline Archive Construction}
We construct the archive using a steady-state MAP-Elites~\cite{mouret2015illuminating} procedure. The search maintains a working population $\mathcal{P}$, an elite archive $\mathcal{M}$, and an experience buffer $\mathcal{B}$. The population supports continual variation, the archive stores niche elites, and the experience buffer stores structured experience extracted from multimodal rollout trajectories for policy refinement.

\subsubsection{Policy Parameterization and Behavior Space}
Each search individual is a planning-policy template $\tau$, represented by a fixed task scaffold together with a small set of editable rule modules. The fixed scaffold specifies the task instruction, action interface, and output format, while the editable modules determine how the agent gathers information, interprets progress, commits to actions, and reacts to repeated failures. We write
\begin{equation}
	\label{eq:encoding}
	\tau = (r_{\mathrm{sense}}, r_{\mathrm{progress}}, r_{\mathrm{act}}, r_{\mathrm{recover}}),
\end{equation}
where $r_{\mathrm{sense}}$, $r_{\mathrm{progress}}$, $r_{\mathrm{act}}$, and $r_{\mathrm{recover}}$ govern sensing behavior, progress interpretation, action commitment, and recovery preference, respectively.

To organize policies by execution style, we assign each template a two-dimensional behavior descriptor
\begin{equation}
	\label{eq:bd}
	\mathrm{BD}(\tau) = \big(b_{\mathrm{int}}(\tau),\, b_{\mathrm{goal}}(\tau)\big),
\end{equation}
computed from rollout statistics. The first coordinate,
\begin{equation}
	\label{eq:bint}
	b_{\mathrm{int}}(\tau)
	=
	\frac{N_{\mathrm{interact}}(\tau)}{N_{\mathrm{total}}(\tau)},
\end{equation}
measures \emph{interaction intensity}, namely the proportion of information-gathering interactions during execution, where $N_{\mathrm{interact}}(\tau)$ denotes the number of interaction-oriented actions and $N_{\mathrm{total}}(\tau)$ denotes the total number of executed actions in the rollout.

The second coordinate,
\begin{equation}
	\label{eq:bgoal}
	b_{\mathrm{goal}}(\tau)
	=
	\frac{N_{\mathrm{adv}}(\tau)}
	{N_{\mathrm{adv}}(\tau)+N_{\mathrm{reg}}(\tau)+N_{\mathrm{rep}}(\tau)+\epsilon},
\end{equation}
measures \emph{goal-directedness}, that is, how consistently the policy converts interaction into forward task progress. Here, $N_{\mathrm{adv}}(\tau)$ is the number of milestone-advancing actions, $N_{\mathrm{reg}}(\tau)$ is the number of regressive or backtracking actions, $N_{\mathrm{rep}}(\tau)$ is the number of repeated low-value actions, and $\epsilon>0$ is a small constant introduced for numerical stability.

The behavior space induced by Eq.~(\ref{eq:bd}) is discretized into a finite set of niches through a niche-mapping operator $\Gamma(\cdot)$. Each niche stores one elite policy in the archive. This design allows the offline search to preserve multiple distinct planning styles instead of retaining only small variations of a single dominant behavior.

\subsubsection{Quality-Diversity Search}
Starting from the initial population, the offline search iteratively generates offspring policies, evaluates their rollout behavior, and updates both the working population and the archive. The search can be viewed as a quality-diversity optimization process in which new policies are generated by differential rule-sentence recombination and probabilistic experience-guided mutation, and are then organized according to their behavior descriptors and quality scores.

At iteration $t$, a base template $\tau_A$ is selected from the archive $\mathcal{M}$ with probability $p_a$ when archive elites are available; otherwise, it is selected from the current population $\mathcal{P}$. Two additional templates, denoted by $\tau_B$ and $\tau_C$, are sampled uniformly from $\mathcal{P}$. We then extract from $\tau_B$ the rule sentences that provide behavioral content relative to $\tau_C$, and denote the resulting set of differential rule sentences by
\begin{equation}
	\label{eq:delta}
	\Delta(\tau_B,\tau_C).
\end{equation}

These differential rule sentences are first revised by a sentence-level mutation operator $\mathcal{M}_s(\cdot)$, yielding
\begin{equation}
	\label{eq:delta_mut}
	\widetilde{\Delta}
	=
	\mathcal{M}_s\!\big(\Delta(\tau_B,\tau_C)\big).
\end{equation}
The revised differential sentences are then inserted into the corresponding rule modules of $\tau_A$. To control inheritance strength, each newly inserted sentence is retained independently with probability $p_r$. Let $\mathcal{R}(\tau_A,\widetilde{\Delta})$ denote this insertion-and-retention operator. The recombined offspring is
\begin{equation}
	\label{eq:recombine}
	\tau^{\mathrm{rec}}
	=
	\mathcal{R}(\tau_A,\widetilde{\Delta}).
\end{equation}

To further increase behavioral diversity and incorporate accumulated search experience, we apply experience-guided mutation to the recombined offspring with probability $p_m$. Let $\mathcal{M}_e(\cdot;\mathcal{B})$ denote the experience-guided mutation operator based on the experience buffer $\mathcal{B}$. The final offspring is
\begin{equation}
	\label{eq:offspring}
	\tau^{\mathrm{off}}
	=
	\begin{cases}
		\mathcal{M}_e\!\left(\tau^{\mathrm{rec}};\mathcal{B}\right), & \text{with probability } p_m,\\[4pt]
		\tau^{\mathrm{rec}}, & \text{otherwise}.
	\end{cases}
\end{equation}
Here, $\mathcal{M}_e$ revises one or more rule sentences according to matched experience patterns retrieved from $\mathcal{B}$. Success patterns reinforce sentence structures associated with effective progress, whereas failure modes weaken, rewrite, or remove sentence structures repeatedly associated with stalled execution.

The resulting offspring $\tau^{\mathrm{off}}$ is evaluated by rollout execution to obtain its behavior descriptor $\mathrm{BD}(\tau^{\mathrm{off}})$ and quality score
\begin{equation}
	\label{eq:quality}
	Q(\tau^{\mathrm{off}})
	=
	\alpha\,\mathrm{SR}(\tau^{\mathrm{off}})
	+
	(1-\alpha)\,\mathrm{Eff}(\tau^{\mathrm{off}}),
\end{equation}
where $\mathrm{SR}(\tau^{\mathrm{off}})\in[0,1]$ denotes task success rate, $\mathrm{Eff}(\tau^{\mathrm{off}})\in[0,1]$ summarizes execution efficiency, and $\alpha\in[0,1]$ balances completion quality and interaction efficiency.

For population update, a comparison subset $\mathcal{S}\subseteq\mathcal{P}$ is sampled, and the offspring replaces the lowest-quality candidate in that subset:
\begin{equation}
	\label{eq:pop_update}
	\tau_{\mathrm{worst}}
	=
	\arg\min_{\tau \in \mathcal{S}} Q(\tau),
	\qquad
	\mathcal{P}
	\leftarrow
	\big(\mathcal{P}\setminus\{\tau_{\mathrm{worst}}\}\big)\cup\{\tau^{\mathrm{off}}\}.
\end{equation}

For archive update, the offspring is assigned to the niche indexed by
\begin{equation}
	\label{eq:niche_assign}
	b=\Gamma\!\big(\mathrm{BD}(\tau^{\mathrm{off}})\big).
\end{equation}
Let $\mathcal{M}[b]$ denote the current elite stored in niche $b$. The archive is updated according to
\begin{equation}
	\label{eq:archive_update}
	\mathcal{M}[b]
	=
	\begin{cases}
		\tau^{\mathrm{off}}, & \text{if niche } b \text{ is empty or } Q(\tau^{\mathrm{off}}) > Q(\mathcal{M}[b]),\\[4pt]
		\mathcal{M}[b], & \text{otherwise}.
	\end{cases}
\end{equation}

Finally, the rollout trajectory of $\tau^{\mathrm{off}}$ is converted into trajectory statistics and structured experience, including success patterns and failure modes, which are added to the experience buffer $\mathcal{B}$. Through this iterative process, the offline stage gradually constructs a compact archive of high-quality yet behaviorally distinct planning-policy templates for later online recovery.

\begin{algorithm}[t]
	\caption{Offline Archive Construction of Planning-Policy Templates}
	\label{alg:offline_qd}
	\small
	\KwIn{Seed template set $\mathcal{S}_0$, population size $N$, archive $\mathcal{M}$, experience buffer $\mathcal{B}$, iteration budget $T$}
	\KwOut{Policy archive $\mathcal{M}$}
	
	Initialize empty archive $\mathcal{M}$ and empty experience buffer $\mathcal{B}$\;
	Generate initial population $\mathcal{P}^{(0)}=\{\tau_1,\dots,\tau_N\}$ from $\mathcal{S}_0$ by editing rule modules\;
	Set current population $\mathcal{P}\leftarrow \mathcal{P}^{(0)}$\;
	
	\For{$t=1$ \KwTo $T$}{
		Select base template $\tau_A$ from $\mathcal{M}$ with probability $p_a$ when archive elites are available; otherwise select $\tau_A$ from $\mathcal{P}$\;
		Sample two templates $\tau_B,\tau_C$ uniformly from $\mathcal{P}$\;
		
		Extract differential rule sentences $\Delta(\tau_B,\tau_C)$\;
		Compute revised differential sentences $\widetilde{\Delta}=\mathcal{M}_s(\Delta(\tau_B,\tau_C))$\;
		Construct recombined offspring $\tau^{\mathrm{rec}}=\mathcal{R}(\tau_A,\widetilde{\Delta})$\;
		
		\eIf{rand() $< p_m$}{
			Generate offspring $\tau^{\mathrm{off}}=\mathcal{M}_e(\tau^{\mathrm{rec}};\mathcal{B})$\;
		}{
			Set $\tau^{\mathrm{off}}\leftarrow \tau^{\mathrm{rec}}$\;
		}
		
		Roll out policy $\pi_{\tau^{\mathrm{off}}}$ to obtain trajectory $\xi$\;
		Compute trajectory statistics and extract structured experience from $\xi$\;
		Compute behavior descriptor $\mathrm{BD}(\tau^{\mathrm{off}})$ using Eqs.~(\ref{eq:bint}) and (\ref{eq:bgoal})\;
		Compute quality score $Q(\tau^{\mathrm{off}})$ using Eq.~(\ref{eq:quality})\;
		
		Sample a comparison subset $\mathcal{S}\subseteq\mathcal{P}$\;
		Let $\tau_{\mathrm{worst}}=\arg\min_{\tau\in\mathcal{S}} Q(\tau)$\;
		Update population
		$\mathcal{P}\leftarrow(\mathcal{P}\setminus\{\tau_{\mathrm{worst}}\})\cup\{\tau^{\mathrm{off}}\}$\;
		
		Assign $\tau^{\mathrm{off}}$ to niche $b=\Gamma(\mathrm{BD}(\tau^{\mathrm{off}}))$\;
		\eIf{niche $b$ is empty}{
			Insert $\tau^{\mathrm{off}}$ into $\mathcal{M}[b]$\;
		}{
			\If{$Q(\tau^{\mathrm{off}})>Q(\mathcal{M}[b])$}{
				Replace the current elite in $\mathcal{M}[b]$ with $\tau^{\mathrm{off}}$\;
			}
		}
		
		Add the extracted structured experience to $\mathcal{B}$\;
	}
	\Return{$\mathcal{M}$}\;
\end{algorithm}

\subsection{Online Adaptive Recovery}
During online execution, the agent follows one policy at a time in a closed loop. However, a locally reasonable policy may still become unproductive over a long horizon, for example by repeatedly interacting with the environment without advancing the task. Our online recovery mechanism addresses this failure mode through three operations: maintaining a recoverable checkpoint, detecting sustained stagnation, and switching to a behaviorally distinct archive policy when necessary.

\subsubsection{Checkpointing and Stall Detection}
To support recovery, the runtime maintains a checkpoint
\begin{equation}
	\label{eq:checkpoint}
	c_t=(\omega_t,h_t,s_t),
\end{equation}
where $\omega_t$ denotes the recoverable environment state maintained by the runtime, $h_t$ is the interaction-history prefix, and $s_t$ is the current progress state. The checkpoint is updated only when execution reaches a stable milestone boundary.

Since behavioral lock-in is not directly observable during execution, we use sustained non-advancement of the progress state as its operational signal. Specifically, the current policy is regarded as stalled if the progress state does not change for $\delta_s$ consecutive steps:
\begin{equation}
	\label{eq:stall}
	D_{\mathrm{stall}}(t)=1
	\iff
	\big[s_{t-\delta_s+1}=\cdots=s_t\big].
\end{equation}
Here, $\delta_s$ is a preset threshold. This criterion captures the practical situation in which the agent continues producing actions but fails to move the task to a new milestone for an extended period.

\subsubsection{Behavior-Aware Policy Switching}
Once a stall is detected, continuing with the same planning policy is often ineffective because the current execution behavior is no longer producing meaningful progress. Instead, the runtime rolls back to the latest checkpoint and switches to another archive policy whose execution style is sufficiently different from that of the current one. The archive therefore serves as a recovery repertoire rather than merely an offline record of diverse solutions.

Policy switching is performed in behavior space rather than directly in policy-parameter space. Since stall is defined as persistent lack of progress under the current execution behavior, effective recovery requires a substantive change in behavior rather than a minor variation of the same strategy. Therefore, switching to a behaviorally distant policy increases the probability of escaping the current stagnation regime.

During each recovery phase, we maintain a set $\mathcal{H}_t$ of archive policies that have already been attempted since the last checkpoint update. This mechanism prevents cyclic switching among a small subset of policies and ensures that each recovery phase explores multiple distinct behavior regions before the checkpoint is updated.

We measure policy difference directly in behavior space. For an archive candidate $\tau$ and the current policy $\tau_{\mathrm{curr}}$, the normalized descriptor-space distance is defined as
\begin{equation}
	\label{eq:distance}
	D_b(\tau,\tau_{\mathrm{curr}})
	=
	\frac{
		\left\|
		\mathrm{BD}(\tau)-\mathrm{BD}(\tau_{\mathrm{curr}})
		\right\|_2
	}{\sqrt{2}},
\end{equation}
where the denominator is the maximum Euclidean distance in the descriptor space $[0,1]^2$, so that $D_b \in [0,1]$.

Given the current stalled policy, the next policy is selected as the archive entry that is farthest from it in behavior space among the candidates not yet tried in the current recovery phase:
\begin{equation}
	\label{eq:switching}
	\tau^*
	=
	\arg\max_{\tau \in \mathcal{M}\setminus \mathcal{H}_t}
	D_b(\tau,\tau_{\mathrm{curr}}).
\end{equation}
This rule promotes a substantive change in execution style rather than a minor variation of the same unproductive behavior.

If all archive entries have been attempted since the last checkpoint update, the tried set is cleared except for the current policy, and selection resumes from the refreshed candidate pool. After selecting $\tau^*$, the runtime resumes planning from checkpoint $c_t$ under the new policy and adds $\tau^*$ to $\mathcal{H}_t$. Whenever a new milestone is reached, the checkpoint is updated and the tried set is reset accordingly. The checkpoint mechanism ensures that policy switching changes future execution strategy without discarding already achieved progress.

Overall, the framework forms a closed loop between offline policy discovery and online execution. The offline stage discovers a diverse set of effective planning-policy templates and organizes them in a behavior-indexed archive, while the online stage monitors execution progress and dynamically switches policies when persistent stall is detected. In this way, behavioral diversity discovered offline is used not only for exploration during search but also as a practical recovery resource during deployment.

\begin{algorithm}[t]
	\caption{Online Adaptive Recovery via Behavior-Aware Policy Switching}
	\label{alg:online_recovery}
	\small
	\KwIn{Current policy $\tau_{\mathrm{curr}}$, policy archive $\mathcal{M}$, current runtime state}
	\KwOut{Updated policy and continued execution}
	
	Initialize checkpoint $c_t$ as the latest milestone-consistent runtime state\;
	Initialize tried-policy set $\mathcal{H}_t \leftarrow \{\tau_{\mathrm{curr}}\}$\;
	
	\While{task not finished}{
		Execute one step under $\tau_{\mathrm{curr}}$\;
		Observe updated history and progress state\;
		
		\If{a new milestone is reached}{
			Update checkpoint $c_t$\;
			Reset $\mathcal{H}_t \leftarrow \{\tau_{\mathrm{curr}}\}$\;
		}
		
		Evaluate $D_{\mathrm{stall}}(t)$ using Eq.~(\ref{eq:stall})\;
		
		\If{$D_{\mathrm{stall}}(t)=1$}{
			Roll back runtime state to checkpoint $c_t$\;
			\If{$\mathcal{M}\setminus \mathcal{H}_t = \emptyset$}{
				Reset $\mathcal{H}_t \leftarrow \{\tau_{\mathrm{curr}}\}$\;
			}
			Select
			$\tau^*=\arg\max_{\tau \in \mathcal{M}\setminus \mathcal{H}_t}
			D_b(\tau,\tau_{\mathrm{curr}})$\;
			Set $\tau_{\mathrm{curr}} \leftarrow \tau^*$\;
			Update $\mathcal{H}_t \leftarrow \mathcal{H}_t \cup \{\tau_{\mathrm{curr}}\}$\;
			Resume execution from checkpoint $c_t$\;
		}
	}
	\Return{final execution under the updated policy}\;
\end{algorithm}

\section{Experiments}
\label{sec:experiments}

\subsection{Experimental Setup}
\label{subsec:experimental-setup}
We evaluate on TDW-MAT~\cite{zhang2023building} (built on TDW~\cite{gan2020threedworld,gan2022threedworld}) using the standard \emph{food} and \emph{stuff} splits.

Our TDW experiments follow the \textbf{single-agent} protocol with episodes capped at \textbf{2000} steps, and use the benchmark carrying rules (at most two target objects per trip without a container, and up to three with a container). For fair comparison, all methods use the same LLM backbone, decoding parameters, and interaction budget where applicable. Our full pipeline uses \textbf{GPT-4} for planning-policy generation by default. The stall trigger is $\delta{=}10$ and the QD archive uses a $10{\times}10$ behavior grid. Further experimental details are provided in the Appendix.

To test whether the recovery mechanism transfers beyond object transport, we also conduct a pilot visual task on VLN-CE~\cite{krantz2020vlnce}, a continuous vision-and-language navigation benchmark built from R2R instructions~\cite{anderson2018r2r} and Matterport3D scenes~\cite{chang2017matterport3d}. We evaluate an 11-scene subset of R2R \texttt{val\_unseen} ($N{=}1839$ episodes).

\paragraph{Baselines and related work.}
\textbf{RHP} (Rule-based Hierarchical Planner) is the strong baseline from the original ThreeDWorld Transport Challenge, as described in \cite{zhang2023building}: a hierarchical planner whose high level follows heuristic rules and whose low level uses an A$^\star$-based planner to navigate on a semantic map, with \emph{frontier exploration} that randomly samples waypoints from unexplored regions as sub-goals.
\textbf{CoELA}~\cite{zhang2023building} (\emph{Cooperative Embodied Agents}) is a modular framework that builds embodied agents from large language models: specialist modules handle perception, high-level planning, and low-level control. Table~\ref{tab:main_results_v2} compares RHP, CoELA, and \textbf{Ours w/ QD} under the single-agent protocol. CoELA and our method are each evaluated with GPT-4, LLaMA-2, and CoLLAMA-2 backbones.

We report both task success and average token consumption to examine inference efficiency. Table~\ref{tab:llm_baselines} provides an additional matched comparison with ReAct, Reflexion, and Tree-of-Thoughts, while Table~\ref{tab:ablation_v2} evaluates the contribution of online switching and archive selection.

\subsection{Main Results}
\label{subsec:main-results}

\begin{table}[t]
\centering
\caption{Single-agent TDW-MAT results under a unified 2000-step cap. Total is the macro-average of Food and Stuff; Token(k) denotes average token consumption in thousands.}
\label{tab:main_results_v2}
\resizebox{\columnwidth}{!}{%
\begin{tabular}{llcccc}
\toprule
Method & Backbone & Food $\uparrow$ & Stuff $\uparrow$ & Total $\uparrow$ & Token(k) $\downarrow$ \\
\midrule
RHP & -- & 0.33 & 0.24 & 0.28 & -- \\
\midrule
\multirow{3}{*}{CoELA}
 & GPT-4 & 0.42 & 0.36 & 0.39 & 69.42 \\
 & LLaMA-2 & 0.35 & 0.24 & 0.30 & 61.50 \\
 & CoLLAMA-2 & 0.37 & 0.31 & 0.34 & 64.88 \\
\midrule
\multirow{3}{*}{\textbf{Ours w/ QD}}
 & GPT-4 & \textbf{0.51} & \textbf{0.43} & \textbf{0.47} & 62.36 \\
 & LLaMA-2 & 0.41 & 0.37 & 0.39 & \textbf{51.21} \\
 & CoLLAMA-2 & 0.44 & 0.39 & 0.41 & 58.77 \\
\bottomrule
\end{tabular}%
}
\end{table}

\begin{table}[t]
\centering
\caption{Direct LLM-planning baselines under the same single-agent GPT-4 protocol and 2000-step budget (Food/Stuff/Total success).}
\label{tab:llm_baselines}
\resizebox{\columnwidth}{!}{%
\begin{tabular}{lccc}
\toprule
Method & Food Success $\uparrow$ & Stuff Success $\uparrow$ & Total Success $\uparrow$ \\
\midrule
ReAct~\cite{yao2022react} & 0.21 & 0.26 & 0.24 \\
Reflexion~\cite{shinn2023reflexion} & 0.28 & 0.26 & 0.27 \\
Tree-of-Thoughts~\cite{yao2023tree} & 0.14 & 0.17 & 0.15 \\
\textbf{Ours (full QD)} & \textbf{0.51} & \textbf{0.43} & \textbf{0.47} \\
\bottomrule
\end{tabular}%
}
\end{table}

Table~\ref{tab:main_results_v2} reports the single-agent TDW-MAT results for RHP, CoELA, and our method. We additionally compare our method with ReAct, Reflexion, and Tree-of-Thoughts under the same GPT-4 planning protocol and 2000-step budget, as shown in Table~\ref{tab:llm_baselines}.

Compared with RHP (2000 steps), \textbf{Ours w/ QD (GPT-4)} improves from \textbf{0.33 to 0.51} on \emph{food} and from \textbf{0.24 to 0.43} on \emph{stuff}. Relative to CoELA with GPT-4 (\textbf{0.42} / \textbf{0.36} on \emph{food} / \emph{stuff}), \textbf{Ours w/ QD (GPT-4)} reaches \textbf{0.51} and \textbf{0.43} under the same step budget. The improvement is consistent across both task categories rather than being driven by a single split. The larger gain on \emph{stuff} is particularly relevant because its longer search and transport sequences leave less room to recover from an early detour.

\textbf{Token vs. success.} Our GPT-4 configuration uses \textbf{62.36k} tokens on average, compared with \textbf{69.42k} for CoELA with GPT-4. The LLaMA-2 configuration requires the fewest tokens among our variants (\textbf{51.21k}), showing that the QD policy can improve success without increasing inference cost. This matters in the MM setting because the final submission requires a concise but complete experimental story: the same policy family should be judged not only by raw success, but also by whether it does so economically under a fixed interaction cap.

More importantly, these gains support our core claim about diversity-driven robustness. Many failures come from repeated local loops rather than one catastrophic mistake. A single fixed prompt-policy tends to preserve the same interaction bias when the environment changes subtly; the QD archive provides behaviorally distinct alternatives so the agent can switch execution style when progress stalls.

\begin{table}[t]
\centering
\caption{Pilot results on VLN-CE R2R \texttt{val\_unseen} (11 MP3D scenes; $N{=}1839$ episodes).}
\label{tab:vln_results}
\begin{tabular}{lccc}
\toprule
Method & SR $\uparrow$ & SPL $\uparrow$ & NE $\downarrow$ \\
\midrule
CMA (frozen) & 35.2\% & 32.9\% & 6.67 \\
\textbf{CMA + QD} & \textbf{37.3\%} & \textbf{34.1\%} & \textbf{6.51} \\
\bottomrule
\end{tabular}
\end{table}

\textbf{Cross-domain visual navigation.} Table~\ref{tab:vln_results} evaluates the same offline-archive and online stall-switching mechanism on VLN-CE. We keep the cross-modal attention (CMA) RGB-D navigation policy~\cite{krantz2020vlnce} frozen: CMA executes routine navigation, while a separate R2R archive briefly supplies a behaviorally complementary recovery policy after a detected stall before control returns to CMA. This archive is constructed from CMA navigation rollouts with stall-recovery descriptors and contains no TDW transport rules. Adding QD improves success rate from 35.2\% to 37.3\% ($+2.1$ points), increases SPL from 32.9\% to 34.1\%, and reduces navigation error from 6.67 to 6.51. Although this pilot is not intended as a VLN state-of-the-art comparison, it provides mechanism-level evidence that diversity-based stall recovery transfers to a distinct visual navigation domain. The pilot further shows that the archive is useful even when the control problem shifts from object transport to waypoint following, which supports the paper’s broader claim that behavioral diversity is a reusable recovery asset rather than a task-specific trick.

\begin{table}[t]
\centering
\caption{Ablation results on online switching and archive selection (Food/Stuff/Total success; 2000 steps). `QD archive' is used in all Ours variants. `OS' denotes online stall-based switching; `ArchSel' denotes the elite selection criterion.}
\label{tab:ablation_v2}
\resizebox{\columnwidth}{!}{%
\begin{tabular}{lccc}
\toprule
Setting & Food Success $\uparrow$ & Stuff Success $\uparrow$ & Total Success $\uparrow$ \\
\midrule
RHP & 0.33 & 0.24 & 0.28 \\
Ours (no OS) & 0.37 & 0.25 & 0.31 \\
Ours (OS + Abl. ArchSel) & 0.44 & 0.39 & 0.41 \\
\textbf{Ours (full QD)} & \textbf{0.51} & \textbf{0.43} & \textbf{0.47} \\
\bottomrule
\end{tabular}%
}
\end{table}

\begin{figure}[t]
    \centering
    \includegraphics[width=0.95\columnwidth]{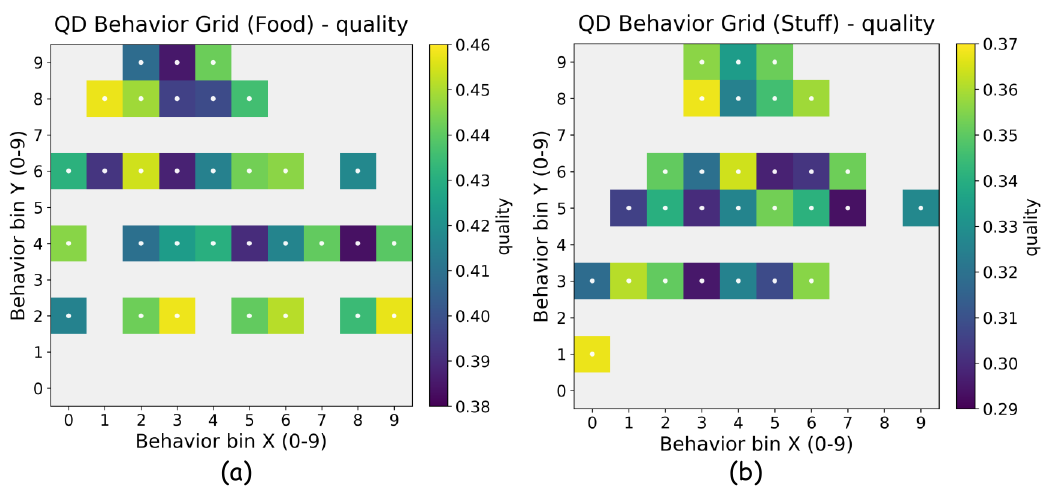}
    \caption{QD behavior-grid visualization (10$\times$10); colored cells indicate occupied elite niches across the interaction-intensity and goal-directedness descriptors.}
    \Description{A ten-by-ten behavior-descriptor grid with colored occupied cells representing elite policies retained by quality-diversity search.}
    \label{fig:qd_behavior_grid}
\end{figure}

Beyond aggregate success and token cost, Fig.~\ref{fig:qd_behavior_grid} visualizes the learned QD repertoire in the behavior-descriptor grid. The broad set of occupied bins indicates that the archive does not collapse to a single prompt-policy style, but instead retains multiple complementary behaviors. This diversity is precisely what enables our online switching mechanism to escape low-yield loops: when progress stalls under one behavior mode, the runtime can select an elite from a different region of the grid and resume progress under the same strict step budget.

These results support the role of behavioral diversity in long-horizon recovery: when execution stalls, the archive provides alternative policies rather than requiring the current policy to continue in the same interaction mode.

\subsection{Ablation Study}
\label{subsec:ablation}
We conduct an ablation study to validate the causal effect of two key components behind our method under the \textbf{2000}-step single-agent protocol. Specifically, we separate: (i) \textbf{stall-based online switching}, which controls whether the agent can change its execution policy when progress stalls; and (ii) \textbf{behavior-indexed archive selection}, which controls how an elite policy is chosen from the offline QD archive at each switching moment.

All variants are evaluated on the same TDW-MAT \emph{food}/\emph{stuff} splits and report success for \emph{food}, \emph{stuff}, and their macro-average \emph{total}. We keep the backbone LLM and the offline QD archive consistent, so the only differences across ablations are the online control logic and the archive selection criterion. This ensures that the observed changes in success can be attributed to the intended design choices rather than training or model capacity variations.

In our full method, the online recovery loop monitors execution progress and triggers a stall event when meaningful progress stops. After stall, the agent performs an online switching step to select a behaviorally complementary elite from the archive, so it can break the current low-yield interaction mode and continue pursuit under the same step budget.

The ablation \textbf{Ours (no OS)} disables stall-triggered switching at execution time. Intuitively, this tests whether a diverse QD archive alone is sufficient: if switching is required to escape time-dependent failure modes, then success should drop when the agent is forced to follow a single fixed policy throughout an episode.

The ablation \textbf{Ours (OS + Abl. ArchSel)} keeps the online switching loop but removes behavior matching from archive selection. Instead of selecting elites that are aligned with the current behavior descriptor, it relies on a simplified criterion (quality-only, or an equivalent random elite choice). This variant tests whether robustness requires not only diverse candidates but also \emph{correct routing} to the right behavior mode at the right time.

Together, these results validate our design rationale: (1) RHP provides a strong base, (2) online switching is essential for long-horizon robustness, and (3) behavior-aware elite selection is required to fully realize the benefit of the offline QD archive.

\begin{figure}[t]
 \centering
 \includegraphics[width=0.98\columnwidth]{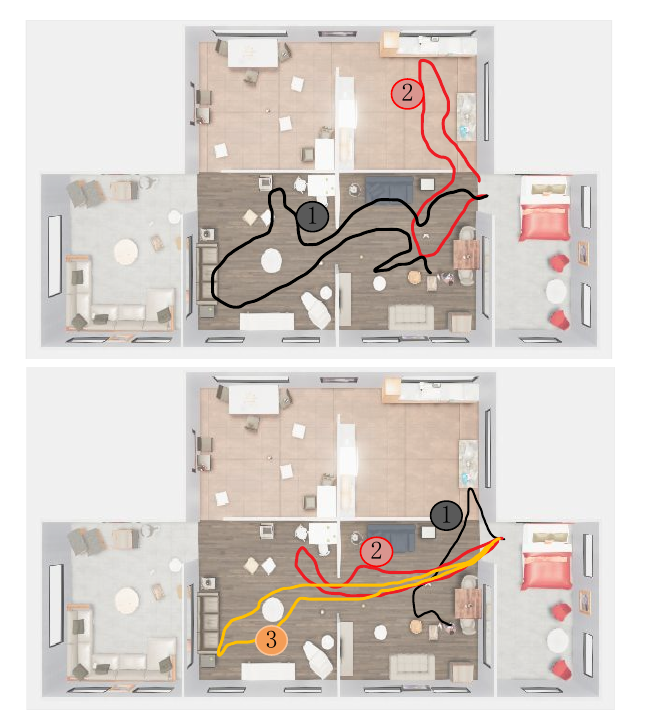}
 \caption{Case-study trajectories in a representative transport episode. \textbf{Top: without QD.} The agent completes two delivery cycles (Route~1 and Route~2) but repeatedly spends a large motion budget on broad exploration and detours before reaching the correct delivery region. \textbf{Bottom: with QD.} After switching to a behaviorally distinct archive policy, the agent prioritizes nearby exploration and acquisition, finishes deliveries in familiar regions first, and only then expands to farther areas, yielding shorter, more structured trajectories and more diverse execution strategies.}
 \Description{Top-down trajectories comparing no-QD and QD-enabled planning in a transport episode, with numbered routes indicating sequential delivery cycles.}
 \label{fig:case_food_transport}
\end{figure}

\subsection{Case Study}
\label{subsec:case-study}

Figure~\ref{fig:case_food_transport} visualizes top-down trajectories for a representative \emph{food} transport episode under a strict 2000-step cap. Under the TDW-MAT single-agent protocol, the agent can transport at most two food objects per trip without a scene container. This constraint makes the motion budget exceptionally tight, as any redundant exploration directly reduces the remaining steps available for subsequent delivery cycles. In this cluttered indoor layout, the agent must navigate between a kitchen/dining area and a bedroom delivery region while avoiding obstacles and repetitive paths.

\textbf{Without QD (top): repeated detours and exploration bias.} In the top panel, the agent follows a fixed planning-policy template throughout the episode. Route~1 shows the first transportation cycle, where the agent adopts a broad exploration routine due to initial environmental uncertainty. It traverses multiple rooms and revisits corridors before finally reaching the bedroom. Route~2 shows the second cycle, where a common long-horizon failure occurs: the policy replays its previous exploration bias instead of exploiting the known map. This leads to a second long, winding trajectory that exhausts the interaction budget, leaving insufficient steps to complete a third delivery. The agent's inability to switch behavioral modes causes it to remain locked in an unproductive exploration style.

\textbf{With QD (bottom): nearby-first strategy and adaptive expansion.} The bottom panel shows the episode with QD-enabled online recovery. When the initial policy shows signs of stalling or inefficiency, the system switches to a behaviorally distinct archive policy. This new policy prioritizes \emph{nearby} rooms and objects, aiming for immediate pickups and deliveries. By completing deliveries in familiar regions first (Route~1 and Route~2) and deferring distant exploration (Route~3), the agent significantly reduces unnecessary navigation. The trajectories are visibly shorter and more direct, allowing the agent to fit an additional delivery cycle within the same 2000-step budget. This adaptive re-routing demonstrates the practical value of behavioral diversity for long-horizon robustness.

\section{Conclusion}
\label{sec:conclusion}
This paper investigates behavioral diversity in long-horizon multimodal embodied planning. We identify behavioral lock-in as a primary failure mode and propose a Quality-Diversity framework that evolves planning-policy templates into a diverse offline archive. During execution, a progress-aware recovery mechanism detects stalls and switches to behaviorally distinct policies, helping the agent escape unproductive interaction loops. Experiments on ThreeDWorld transport and a pilot VLN-CE visual navigation task show improved task success under long-horizon interaction constraints. Overall, the results indicate that combining offline diversity discovery with online behavioral adaptation is a promising direction for robust embodied planning.

The central lesson is that robustness does not come from making one policy increasingly elaborate. It comes from preserving several useful ways of acting and choosing among them when the current one stops matching the situation. The transport results show this effect under a tight interaction budget, while the visual-navigation pilot suggests that the same recovery principle is not tied to object manipulation. Together, these findings motivate future work on larger visual benchmarks, learned stall detectors, and archives whose behavior descriptors are grounded in task-relevant navigation outcomes rather than prompt statistics alone.

\clearpage
\begin{acks}
This work was supported by the Natural Science Foundation of Jiangsu Province (Grant No.\ BK20230419).
\end{acks}

\bibliographystyle{ACM-Reference-Format}
\balance
\bibliography{acmmm_work_refs}

@article{lehman2011abandoning,
  title={Abandoning objectives: Evolution through the search for novelty alone},
  author={Lehman, Joel and Stanley, Kenneth O},
  journal={Evolutionary computation},
  volume={19},
  number={2},
  pages={189--223},
  year={2011},
  publisher={MIT Press}
}

@article{pugh2016qualitydiversity,
  title={Quality diversity: A new frontier for evolutionary computation},
  author={Pugh, Justin K and Soros, Lisa B and Stanley, Kenneth O},
  journal={Frontiers in Robotics and AI},
  volume={3},
  pages={40},
  year={2016},
  publisher={Frontiers Media SA}
}

@inproceedings{fernando2024promptbreeder,
  title={Promptbreeder: Self-Referential Self-Improvement via Prompt Evolution},
  author={Fernando, Chrisantha and Banarse, Dylan Sunil and Michalewski, Henryk and Osindero, Simon and Rockt{\"a}schel, Tim},
  booktitle={International Conference on Machine Learning},
  pages={13481--13544},
  year={2024},
  organization={PMLR}
}

@inproceedings{huang2023inner,
  title={Inner Monologue: Embodied Reasoning through Planning with Language Models},
  author={Huang, Wenlong and Xia, Fei and Xiao, Ted and Chan, Harris and Liang, Jacky and Florence, Pete and Zeng, Andy and Tompson, Jonathan and Mordatch, Igor and Chebotar, Yevgen and others},
  booktitle={Conference on Robot Learning},
  pages={1769--1782},
  year={2023},
  organization={PMLR}
}

@inproceedings{song2023llm_planner,
  title={{LLM-Planner}: Few-Shot Grounded Planning for Embodied Agents with Large Language Models},
  author={Song, Chan Hee and Wu, Jiaman and Washington, Clayton and Sadler, Brian M and Chao, Wei-Lun and Su, Yu},
  booktitle={Proceedings of the IEEE/CVF international conference on computer vision},
  pages={2998--3009},
  year={2023}
}

@inproceedings{shridharalfworld,
  title={{ALFWorld}: Aligning Text and Embodied Environments for Interactive Learning},
  author={Shridhar, Mohit and Yuan, Xingdi and Cote, Marc-Alexandre and Bisk, Yonatan and Trischler, Adam and Hausknecht, Matthew},
  booktitle={International Conference on Learning Representations},
  year={2021},
  url={https://openreview.net/forum?id=0IOX0YcCdTn}
}

@inproceedings{zitkovich2023rt,
  title={{RT-2}: Vision-Language-Action Models Transfer Web Knowledge to Robotic Control},
  author={Zitkovich, Brianna and Yu, Tianhe and Xu, Sichun and Xu, Peng and Xiao, Ted and Xia, Fei and Wu, Jialin and Wohlhart, Paul and Welker, Stefan and Wahid, Ayzaan and others},
  booktitle={Conference on Robot Learning},
  pages={2165--2183},
  year={2023},
  organization={PMLR}
}

@article{ahn2022can,
  title={{Do As I Can, Not As I Say}: Grounding Language in Robotic Affordances},
  author={Ahn, Michael and Brohan, Anthony and Brown, Noah and Chebotar, Yevgen and Cortes, Omar and David, Byron and Finn, Chelsea and Fu, Chuyuan and Gopalakrishnan, Keerthana and Hausman, Karol and others},
  journal={arXiv preprint arXiv:2204.01691},
  year={2022}
}

@inproceedings{gao2025building,
  title={Building Embodied {EvoAgent}: A Brain-Inspired Paradigm for Bridging Multimodal Large Models and World Models},
  author={Gao, Junyu and Yao, Xuan and Rui, Yong and Xu, Changsheng},
  booktitle={Proceedings of the 33rd ACM International Conference on Multimedia},
  pages={3280--3289},
  year={2025}
}

@article{chen2025exploring,
  title={Exploring embodied multimodal large models: Development, datasets, and future directions},
  author={Chen, Shoubin and Wu, Zehao and Zhang, Kai and Li, Chunyu and Zhang, Baiyang and Ma, Fei and Yu, Fei Richard and Li, Qingquan},
  journal={Information Fusion},
  volume={122},
  pages={103198},
  year={2025},
  publisher={Elsevier}
}

@inproceedings{krantz2020vlnce,
  title={Beyond the Nav-Graph: Vision-and-Language Navigation in Continuous Environments},
  author={Krantz, Jacob and Wijmans, Erik and Majumdar, Arjun and Batra, Dhruv and Lee, Stefan},
  booktitle={European Conference on Computer Vision},
  pages={104--120},
  year={2020},
  publisher={Springer}
}

@inproceedings{anderson2018r2r,
  title={Vision-and-Language Navigation: Interpreting Visually-Grounded Navigation Instructions in Real Environments},
  author={Anderson, Peter and Wu, Qi and Teney, Damien and Bruce, Jake and Johnson, Mark and S{"o}nderby, Soren and Reid, Ian and van den Hengel, Anton},
  booktitle={Proceedings of the IEEE Conference on Computer Vision and Pattern Recognition},
  pages={3674--3683},
  year={2018}
}

@inproceedings{chang2017matterport3d,
  title={Matterport3D: Learning from RGB-D Data in Indoor Environments},
  author={Chang, Angel and Dai, Angela and Funkhouser, Thomas and Halber, Maciej and Niessner, Matthias and Savva, Manolis and Song, Shuran and Zeng, Andy and Zhang, Yinda},
  booktitle={2017 International Conference on 3D Vision},
  pages={667--676},
  year={2017},
  organization={IEEE}
}

@inproceedings{gao2024fast,
  title={Fast-slow test-time adaptation for online vision-and-language navigation},
  author={Gao, Junyu and Yao, Xuan and Xu, Changsheng},
  booktitle={Proceedings of the 41st International Conference on Machine Learning},
  pages={14902--14919},
  year={2024}
}

@inproceedings{yao2022react,
  title={React: Synergizing reasoning and acting in language models},
  author={Yao, Shunyu and Zhao, Jeffrey and Yu, Dian and Du, Nan and Shafran, Izhak and Narasimhan, Karthik R and Cao, Yuan},
  booktitle={The eleventh international conference on learning representations},
  year={2022}
}

@article{yao2023tree,
  title={Tree of thoughts: Deliberate problem solving with large language models},
  author={Yao, Shunyu and Yu, Dian and Zhao, Jeffrey and Shafran, Izhak and Griffiths, Tom and Cao, Yuan and Narasimhan, Karthik},
  journal={Advances in neural information processing systems},
  volume={36},
  pages={11809--11822},
  year={2023}
}

@article{shinn2023reflexion,
  title={Reflexion: Language agents with verbal reinforcement learning},
  author={Shinn, Noah and Cassano, Federico and Gopinath, Ashwin and Narasimhan, Karthik and Yao, Shunyu},
  journal={Advances in neural information processing systems},
  volume={36},
  pages={8634--8652},
  year={2023}
}

@article{zhang2023building,
  title={Building cooperative embodied agents modularly with large language models},
  author={Zhang, Hongxin and Du, Weihua and Shan, Jiaming and Zhou, Qinhong and Du, Yilun and Tenenbaum, Joshua B and Shu, Tianmin and Gan, Chuang},
  journal={arXiv preprint arXiv:2307.02485},
  year={2023}
}

@article{mouret2015illuminating,
  title={Illuminating search spaces by mapping elites},
  author={Mouret, Jean-Baptiste and Clune, Jeff},
  journal={arXiv preprint arXiv:1504.04909},
  year={2015}
}

@article{gan2020threedworld,
  title={{ThreeDWorld}: A Platform for Interactive Multi-Modal Physical Simulation},
  author={Gan, Chuang and Schwartz, Jeremy and Alter, Seth and Mrowca, Damian and Schrimpf, Martin and Traer, James and De Freitas, Julian and Kubilius, Jonas and Bhandwaldar, Abhishek and Haber, Nick and others},
  journal={arXiv preprint arXiv:2007.04954},
  year={2020}
}

@inproceedings{gan2022threedworld,
  title={The {ThreeDWorld} Transport Challenge: A Visually Guided Task-and-Motion Planning Benchmark Towards Physically Realistic Embodied {AI}},
  author={Gan, Chuang and Zhou, Siyuan and Schwartz, Jeremy and Alter, Seth and Bhandwaldar, Abhishek and Gutfreund, Dan and Yamins, Daniel LK and DiCarlo, James J and McDermott, Josh and Torralba, Antonio and others},
  booktitle={2022 International conference on robotics and automation (ICRA)},
  pages={8847--8854},
  year={2022},
  organization={IEEE}
}

@inproceedings{guo2023evoprompt,
  title={{EvoPrompt}: Connecting {LLMs} with Evolutionary Algorithms Yields Powerful Prompt Optimizers},
  author={Guo, Qingyan and Wang, Rui and Guo, Junliang and Li, Bei and Song, Kaitao and Tan, Xu and Liu, Guoqing and Bian, Jiang and Yang, Yujiu},
  booktitle={The Twelfth International Conference on Learning Representations},
  year={2024},
  url={https://openreview.net/forum?id=ZG3RaNIsO8}
}
\end{document}